\documentclass{article}

\usepackage{microtype}
\usepackage{graphicx}
\usepackage{subcaption}
\usepackage{booktabs}
\usepackage{hyperref}

\usepackage[accepted]{icml2026}

\usepackage[utf8]{inputenc}
\usepackage[T1]{fontenc}
\usepackage{amsmath}
\usepackage{amssymb}
\usepackage{mathtools}
\usepackage{xcolor}
\usepackage{array}
\usepackage{float}
\usepackage{enumitem}

\definecolor{relblue}{RGB}{31,119,180}
\definecolor{entred}{RGB}{214,39,40}
\definecolor{darkgray}{RGB}{60,60,60}
\definecolor{tblhdr}{RGB}{31,119,180}
\definecolor{tblalt}{RGB}{234,243,251}

\newcolumntype{C}[1]{>{\centering\arraybackslash}p{#1}}
\newcolumntype{L}[1]{>{\raggedright\arraybackslash}p{#1}}

\newcommand{\relc}[1]{\textcolor{relblue}{#1}}
\newcommand{\entc}[1]{\textcolor{entred}{#1}}

\icmltitlerunning{Relation Before Entity}

\begin{document}

\twocolumn[
\icmltitle{Relation Before Entity: Deferred Commitment\\
in Language Model Factual Recall}

\begin{icmlauthorlist}
  \icmlauthor{Divyansh Agarwal}{uq}
\end{icmlauthorlist}

\icmlaffiliation{uq}{School of Electrical Engineering and Computer Science, The University of Queensland, Brisbane, Australia}
\icmlcorrespondingauthor{Divyansh Agarwal}{d.agarwal@student.uq.edu.au}

\icmlkeywords{Mechanistic Interpretability, Activation Patching, Factual Recall, Language Models, Causal Interventions, Representation Analysis}

\vskip 0.3in
]

\printAffiliationsAndNotice{}

\begin{abstract}
We ask whether relation-type information
(e.g., \textit{capital-of}) and entity-specific information
(e.g., \textit{France}$\to$\textit{Paris}) become
causally active at the final-token position at the same
depth during recall. Using four complementary causal
diagnostics across four decoder-only models and eight prompt
families, we find a robust temporal asymmetry:
\textbf{relation information becomes generation-controlling before
entity information does}. Relation onset precedes entity
onset by $10$--$16$ tested layers ($31$--$44\%$ of network depth)
at threshold $0.4$, with the ordering holding across all $16$
model-threshold combinations for thresholds $0.2$--$0.5$.
Critically, entity information is \emph{not} absent early:
entity-token patching succeeds at $90$--$100\%$ in early layers.
Instead, entity commitment to generation is \textbf{deferred}:
entity information is available at the entity-token position but
becomes generation-controlling at the final token only after being
routed there.
\end{abstract}

\section{Introduction}
\label{sec:intro}
When a model completes ``\textit{The capital of France
is}~\underline{\phantom{x}}'', does it retrieve
\textit{Paris} directly, or does it first settle the
relation \textit{capital-of} at the generation position
before committing to the entity-specific answer? Prior
work has studied \emph{where} factual associations are
stored~\citep{meng2022locating,meng2022mass} and how
subject-token representations are enriched with
attributes~\citep{geva2023dissecting}. \citet{geva2023dissecting}
identify a three-stage information flow in factual recall
and observe that relation and subject information reach
the final token at different points, but do not quantify
a layer-wise onset gap across architectures, use direct
causal transfer measurement, or separate early entity
availability from later final-token commitment. We study
the complementary \emph{temporal} question with direct
causal measurement: when do relation and entity signals
each become generation-controlling at the final token?

We use \emph{entity} broadly to mean the input item supplied
to the relation: a country in factual prompts, a verb in tense
prompts, a noun in plural prompts, an adjective in lexical prompts,
or an element in symbolic prompts. We refer to the prompt position
containing this item as the entity-token position.

\paragraph{Core claim.}
Recall in these controlled prompt families is
\textbf{temporally factorized}. Relation information becomes
causally active at the final-token position in middle layers,
whereas entity commitment is deferred to late layers. This
staging holds across four architectures, eight prompt families
spanning factual, morphological, lexical, and symbolic
transformations, and thresholds $0.2$--$0.5$.

\paragraph{Why this matters.}
If relation and entity information become generation-controlling
at different depths, then (1) layer-targeted interventions such as
model editing or steering may affect relation-type and
answer-identity computation differently, (2) monitoring methods
should distinguish information \emph{availability} from causal
\emph{commitment}, and (3) similar staging may also matter for
chain-of-thought reasoning and multi-hop settings.

\paragraph{Contributions.}
\begin{enumerate}[leftmargin=1.3em,itemsep=0pt,
                  topsep=2pt,parsep=0pt]
  \item Transfer-curve patching shows relation onset
    precedes entity onset by $10$--$16$ tested layers at
    the primary threshold $0.4$, with the ordering robust
    across thresholds $0.2$--$0.5$ (16/16
    model-threshold combinations; \S\ref{sec:exp1}).
  \item Both-change competition pits relation and entity
    signals directly, showing dominance transitions at the
    predicted layers (\S\ref{sec:exp2}).
  \item Entity-token patching shows entity information is
      \emph{available} early at the entity-token position but
      \emph{committed} late at the final-token position, ruling out
      the alternative that entity information is simply absent
      (\S\ref{sec:exp3}).
  \item Steering with relation and entity directions
    supports the asymmetry via a complementary intervention
    (\S\ref{sec:exp4}).
\end{enumerate}

\section{Setup}
\label{sec:setup}

\paragraph{Models.}
Llama-3.2-3B (28 layers), Llama-3-8B (32L),
Qwen2.5-3B (36L), Phi-2 (32L).

\paragraph{Prompt families.}
Eight fill-in-the-blank families: \textit{capital},
\textit{language}, \textit{past tense},
\textit{present participle}, \textit{plural},
\textit{opposite}, \textit{comparative},
\textit{chemical symbol} (see Appendix~\ref{app:families}).
Prompt items are programmatically defined; we audit greedy
generation and first-answer-token statistics as a sanity check.

\paragraph{Terminology.}
\textit{Relation}: the transformation or fact type being
requested (capital-of, past-tense, opposite).
\textit{Entity}: the input item supplied to the relation
and determining the answer (e.g., France, \textit{walk},
\textit{cat}, or gold).

\paragraph{Activation patching.}
At tested layer $\ell$, we replace the donor prompt's
final-token hidden state into the receiver's final-token
position and measure whether the output switches to the
donor's answer. We test every other layer.

\section{Experiment 1: Transfer Curves and Onset}
\label{sec:exp1}

\paragraph{Design.}
\textbf{Relation transfer}: donor and receiver share
entity, differ in relation. Receiver
\textit{capital(France)}, donor \textit{language(France)};
success = \textit{French}. Donor relation applied to
receiver entity tests whether the final-token state is
controlled by relation-type information.
\textbf{Entity transfer}: same relation, different entity.
Receiver \textit{capital(France)}, donor
\textit{capital(Japan)}; success = \textit{Tokyo}. Tests
whether the final-token state is controlled by the
specific answer information.

We report \emph{pair-balanced} relation curves (equal
weighting across six relation pairs) and
\emph{family-balanced} entity curves (equal weighting
across eight families), preventing record-count
weighting artifacts.

\textbf{Onset} is the first tested layer exceeding 0.4
for two consecutive tested layers, capturing stable
causal commitment while avoiding single-layer spikes.
Threshold sensitivity ($0.2$--$0.5$) is in
Appendix~\ref{app:threshold}.

\begin{figure}[t]
  \centering
  \includegraphics[width=\linewidth]{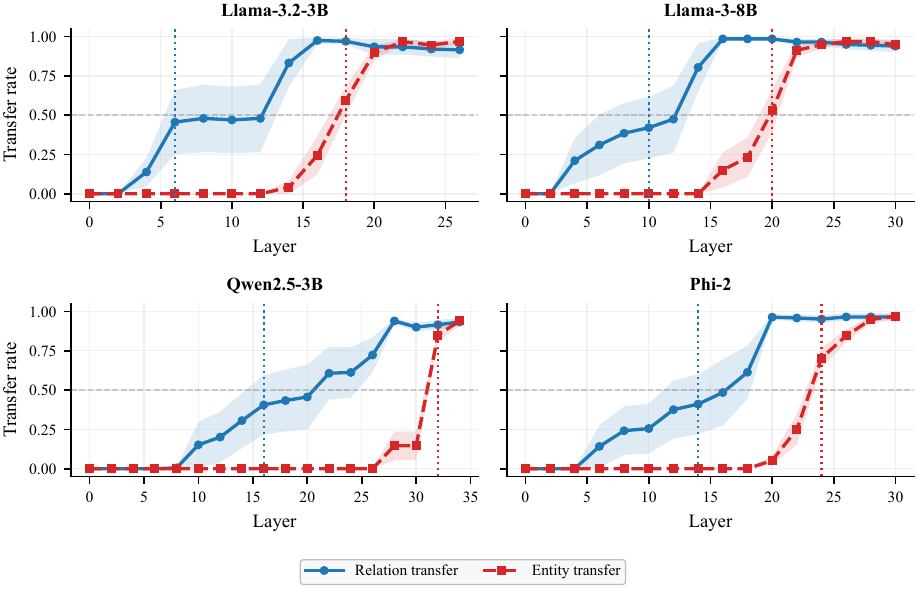}
  \caption{\textbf{Transfer curves.} \relc{Relation
    transfer (blue)} rises before \entc{entity transfer
    (red)} in every model. Entity transfer remains at or
    near 0\% through multiple consecutive layers while
    relation transfer already exceeds $0.4$--$0.8$. Shading
    = SEM across relation pairs or entity families. Dotted lines = onset layers.}
  \label{fig:transfer}
\end{figure}

\paragraph{Results.}
Figure~\ref{fig:transfer} shows relation transfer rising before entity transfer in every model: entity transfer remains near 0\% for multiple layers while relation transfer already exceeds $0.4$--$0.8$. Table~\ref{tab:onset} reports pair-balanced onset layers at threshold $0.4$.

\begin{table}[t]
\centering
\caption{Pair-balanced onset at threshold $0.4$. Relation
  onset precedes entity onset in all models. Ordering
  holds at all thresholds $0.2$--$0.5$ ($16/16$).}
\label{tab:onset}
\vspace{2pt}
\setlength{\tabcolsep}{3.5pt}
\small\renewcommand{\arraystretch}{1.12}
\begin{tabular}{lcccc}
\toprule
Model & Rel.\ onset & Ent.\ onset & Gap & Depth \\
\midrule
Llama-3.2-3B & $L6$  & $L18$ & $12$ & $43\%$ \\
Llama-3-8B   & $L10$ & $L20$ & $10$ & $31\%$ \\
Qwen2.5-3B   & $L16$ & $L32$ & $16$ & $44\%$ \\
Phi-2        & $L14$ & $L24$ & $10$ & $31\%$ \\
\bottomrule
\end{tabular}
\end{table}

\paragraph{Ruling out answer copying.}
A \emph{wrong-entity} control patches a different entity
of the same donor family
(e.g., donor \textit{language(Japan)} into receiver
\textit{capital(France)}). This separates relation-only
transfer from donor-answer copying: output \textit{French}
indicates the donor relation applied to the receiver entity,
whereas output \textit{Japanese} would indicate copying the
donor's specific answer. At the peak relation-only transfer
layer for each model, relation-only transfer reaches
$0.79$--$0.86$ while donor-answer copying remains $\leq 0.11$;
donor-answer copying rises only in later layers once entity
commitment takes over. This indicates that mid-layer patches
carry relational structure rather than simply copying specific
donor answers. The full
wrong-entity control table is provided in
Appendix~\ref{app:controls}.

\section{Experiment 2: Both-Change Competition}
\label{sec:exp2}

\paragraph{Design.}
Donor and receiver differ in \emph{both} relation and
entity, forcing direct competition. Receiver
\textit{capital(France)}, donor
\textit{language(Japan)}. Outputs are classified as:
\textit{original retained} (Paris),
\relc{\textit{relation wins}} (French---donor relation
applied to receiver entity),
\entc{\textit{entity wins}} (Japanese---the donor's answer),
or \textit{mixed}. To our knowledge, this provides a direct
way to measure which signal dominates when relation and entity
information are placed in causal conflict across depth.

\paragraph{Results.}
Relation wins dominate middle layers and entity wins
dominate late layers across all four models
(Figure~\ref{fig:both}; Appendix~\ref{app:bothfig}).
Table~\ref{tab:crossover} shows that the crossover layer
closely tracks the entity onset from Experiment~1: the layer
where entity transfer becomes active is also where entity begins
winning under direct competition.

\begin{table}[t]
\centering
\caption{Both-change crossover and peak layers. Crossover
  aligns with Experiment~1 entity onset within $\leq 2$
  tested layers in every model.}
\label{tab:crossover}
\vspace{2pt}
\setlength{\tabcolsep}{3pt}
\small\renewcommand{\arraystretch}{1.12}
\begin{tabular}{lcccc}
\toprule
Model & Rel.\ peak & Cross. & Ent.\ peak & Ent.\ onset \\
\midrule
Llama-3.2-3B & $L14$ & $L18$ & $L26$ & $L18$ \\
Llama-3-8B   & $L14$ & $L18$ & $L24$ & $L20$ \\
Qwen2.5-3B   & $L26$ & $L32$ & $L34$ & $L32$ \\
Phi-2        & $L20$ & $L24$ & $L28$ & $L24$ \\
\bottomrule
\end{tabular}
\end{table}

\paragraph{Controls.}
Noise-patch controls (random Gaussian vectors) yield
near-zero structured wins ($\leq 0.008$), ruling out
generic perturbation as explanation. Self-patch controls
confirm hook stability ($\geq 99.2\%$ original retained).
Alternate-donor controls show the same qualitative pattern,
supporting robustness to donor choice. Unrelated-donor
diagnostics reveal high late entity-like overwrite but
relation-wins remain near zero ($\leq 0.030$), ruling out
generic patching as explanation for mid-layer relation
dominance. Full control diagnostics are provided in
Appendix~\ref{app:controls}.

\section{Experiment 3: Entity-Token vs.\ Final-Token Patching}
\label{sec:exp3}

\paragraph{Motivation.}
A key alternative explanation is that entity information is
simply absent until late layers, making the delay trivial.
Experiment~3 directly tests this.

\paragraph{Design.}
For same-relation/different-entity pairs, we compare:
(a)~patching the donor's \emph{final-token} state into
the receiver's final-token position, and (b)~patching
the donor's \emph{entity-token} state into the receiver's
entity-token position (83 records, eight families). Here,
the entity-token position is the prompt position containing
the input entity, corresponding to the subject-token position
in factual prompts. If entity information is present early at
this position, entity-token patching should succeed early even
when final-token patching fails.

\begin{figure}[t]
  \centering
  \includegraphics[width=\linewidth]{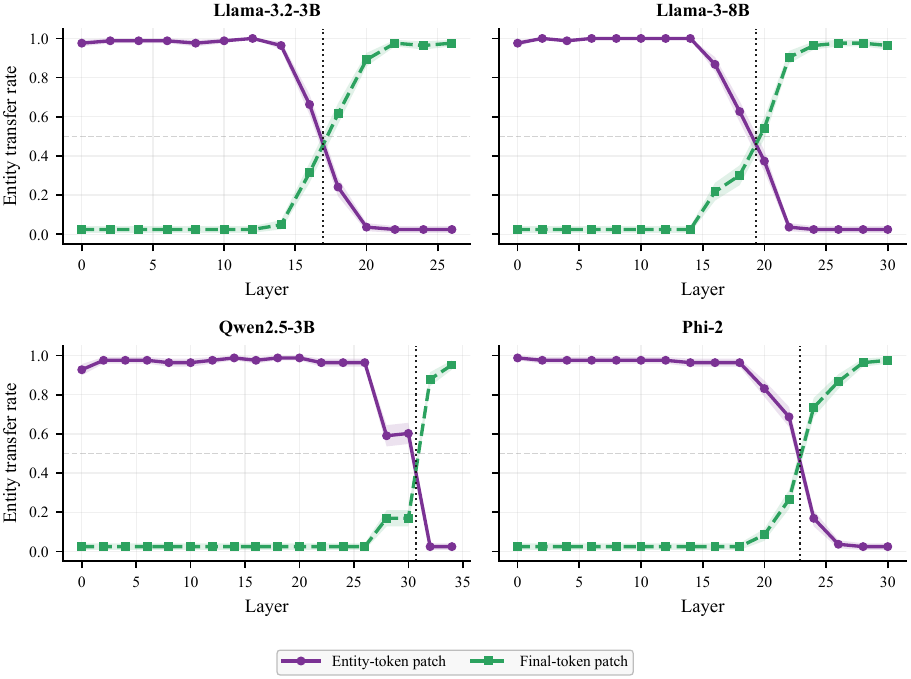}
  \caption{\textbf{Entity-token vs.\ final-token patching.}
  Entity-token patch (purple) transfers entity identity
  at $90$--$100\%$ in early layers. Final-token patch
  (green) is near zero early. The pattern reverses
  late. Entity information exists early but reaches
  the final token only later. Dotted vertical line marks
  the switch layer.}
  \label{fig:subj}
\end{figure}

\paragraph{Results.}
Figure~\ref{fig:subj} shows a clean double dissociation.
Entity-token patching succeeds at \textbf{$90$--$100\%$}
in early and middle layers while final-token patching
remains $\approx 2.4\%$. In late layers the pattern
reverses: final-token patching rises to $88$--$97\%$ and
entity-token patching collapses to $2$--$4\%$.
Table~\ref{tab:switch} shows the crossover matches
entity onset from Experiments~1--2 in every model.

\begin{table}[t]
\centering
\caption{Entity-token/final-token crossover versus Experiment~1
  entity onset. Three complementary diagnostics converge on
  the same transition layers in every model.}
\label{tab:switch}
\vspace{2pt}
\setlength{\tabcolsep}{4pt}
\small\renewcommand{\arraystretch}{1.12}
\begin{tabular}{lcc}
\toprule
Model & Switch layer & Ent.\ onset \\
\midrule
Llama-3.2-3B & $L18$--$L20$ & $L18$ \\
Llama-3-8B   & $L20$--$L22$ & $L20$ \\
Qwen2.5-3B   & $L32$        & $L32$ \\
Phi-2        & $L24$--$L26$ & $L24$ \\
\bottomrule
\end{tabular}
\end{table}

\noindent The alternative explanation is ruled out:
entity information is not absent early. It is available at
the entity-token position from the earliest layers, but
becomes generation-controlling at the final token only after
being routed there. \textit{Deferred entity commitment is
delayed routing, not delayed knowledge.}

\section{Experiment 4: Steering Temporal Asymmetry}
\label{sec:exp4}

As an independent check, we construct relation steering
directions (mean-difference vectors across
same-entity/different-relation prompts) and entity
steering directions (pair-specific
same-relation/different-entity pairs), and apply them
at mid and late layer zones (Appendix~\ref{app:steer}).
Random-direction baselines are $\leq 0.015$,
supporting direction-specificity. Relation directions
steer effectively in middle layers; entity directions
are substantially weaker mid-layer and strongest late.
Relation directions also remain effective late but
are outcompeted by entity commitment under direct
conflict (Experiment~2); relation information
persists but does not win.

\section{Related Work}
\label{sec:related}

\paragraph{Factual recall and model editing.}
\citet{meng2022locating} use causal tracing to identify
MLP computations at subject-token positions that mediate
factual recall, enabling single-fact weight editing.
\citet{meng2022mass} extend this line of work to mass
editing by distributing updates across a range of critical
MLP layers. Our entity-token patching is consistent with
this broad picture: entity information can be causally
available at the token containing the input entity before
it becomes generation-controlling at the final token. These
model-editing results are plausibly related to the
availability--commitment gap we measure, although we do
not directly evaluate weight editing here.

\paragraph{Staged information flow in factual recall.}
\citet{geva2023dissecting} identify a three-step mechanism
for factual attribute extraction: subject enrichment at early
MLP layers, relation propagation to the final token, and
attribute extraction via attention heads. They find that
relation information reaches the final token before subject
information, using attention-edge interventions. Our work
complements this result with direct causal transfer measurements
across four architectures and eight prompt families. We introduce
a both-change competition that places relation and entity signals
in direct conflict, and an entity-token versus final-token patching
analysis showing that entity information can be available early
at the entity-token position while becoming generation-controlling
at the final token only later.

\paragraph{Relation and task representations.}
\citet{todd2024function} show that attention heads transport
compact, causally effective task representations (function
vectors) that generalize beyond the demonstrating context.
\citet{hernandez2024linearity} show that relation-decoding
computations are well approximated by linear maps on subject
representations. \citet{popovivc2026tracing} find that
per-head attention contributions are comparatively strong
features for linear relation classification. These works
support the view that relation and task level information
can be linearly represented inside transformer models. Our
question is complementary: when does relation-level
information become generation-controlling relative to
entity-specific information at the final prediction position?

\paragraph{Causal intervention methods.}
Our methodology follows causal mediation analysis, activation
patching, and transformer-circuits style analysis.
\citet{vig2020investigating} emphasize interventions on
internal model components to distinguish information that is
merely decodable from information that causally mediates
behavior. \citet{heimersheim2024use} offer a systematic treatment 
of activation patching methodology, including how patching 
experiments should be applied and interpreted. \citet{wang2022interpretability} provide a
detailed circuit-level analysis of indirect object
identification in GPT-2 Small. In contrast, we do not claim
to identify a complete circuit; instead, we use causal
interventions to identify a cross-model, cross-family timing
asymmetry in when relation and entity information become
generation-controlling.

\section{Discussion}
\label{sec:discussion}

\paragraph{A four-stage picture.}
Together, the results support a four-stage picture:
\textit{entity-token availability}
$\rightarrow$
\textit{relation becomes final-token active}
$\rightarrow$
\textit{entity becomes final-token committed}
$\rightarrow$
\textit{late overwrite sensitivity}.

Early layers contain entity information at the entity-token
position ($90$--$100\%$ patching success), but this information
is not yet generation-controlling at the final token
($\approx 2.4\%$ transfer). Middle layers make relation information
final-token active; late layers route and commit entity information
to the final token; and the latest layers become broadly
overwrite-sensitive, with unrelated donors inducing late
entity-like overwrite while unrelated relation-wins remain near zero
(Appendix~\ref{app:controls}).

\paragraph{Implications.}
The availability-commitment gap suggests that monitoring
and intervention methods should distinguish
\emph{information being represented} from
\emph{information controlling generation}. A feature may
be decodable at one position or layer without yet causally
determining the next token. In practical terms, middle-layer
interventions may preferentially affect which relation type
is applied (e.g., redirecting from \textit{capital-of} to
\textit{language-of}), whereas late-layer interventions may
preferentially affect which specific answer is produced
(e.g., changing \textit{Paris} to \textit{Tokyo}). This
suggests that single depth monitoring tools may conflate
relation-level and entity-level commitment, especially if
they measure representational availability rather than
causal control. For model editing methods such as
ROME~\citep{meng2022locating}, our results raise the
hypothesis that edits at different depths may target
different stages of recall: relation selection, entity
routing, or final answer commitment. We do not test weight
editing directly, but the observed transition zone provides
a concrete target for future editing and steering analyses.
Whether analogous staging appears in multi-hop reasoning,
where models may need to resolve multiple relation-entity
stages sequentially, remains important future work.

\paragraph{Limitations.}
Our experiments use controlled fill-in-the-blank prompt families
and greedy first-answer generation; natural-language QA, longer
contexts, and free-form reasoning remain untested. The evaluated
models are open-weight decoder-only models in the $3$B--$8$B range,
so exact layer numbers should not be interpreted as universal.
We identify a robust causal timing asymmetry, not a complete circuit:
the mechanisms routing entity information from entity-token positions
to final-token commitment remain to be localized. The central claim is
the ordering (relation before entity commitment), not any single
absolute layer index.

\section{Conclusion}
Recall in controlled prompt families is temporally factorized:
relation information becomes generation-controlling at the final-token
position before entity information does. At threshold $0.4$, relation
onset precedes entity onset by $10$--$16$ tested layers ($31$--$44\%$
of network depth), with the ordering holding across all $16$
model-threshold combinations for thresholds $0.2$--$0.5$. Entity
information is not absent early: it is available at the entity-token
position from the earliest layers, but its commitment to generation is
deferred until routed to the final token. Four complementary causal
diagnostics converge on the same transition layers, providing evidence
for \textbf{deferred entity commitment} in decoder-only transformers.

\section*{Impact Statement}
This work aims to improve mechanistic understanding of recall in
language models. Better understanding may support safety monitoring,
steering, and model editing; we do not identify specific harms requiring
further discussion.

\section*{Acknowledgements}
The author thanks the School of Electrical Engineering and
Computer Science at the University of Queensland for
computational resources, and the anonymous reviewers for
constructive feedback.

\bibliographystyle{icml2026}
\bibliography{references}

\newpage
\appendix
\onecolumn

\section{Both-Change Competition Figure}
\label{app:bothfig}

\begin{figure}[h]
  \centering
  \includegraphics[width=0.85\textwidth]{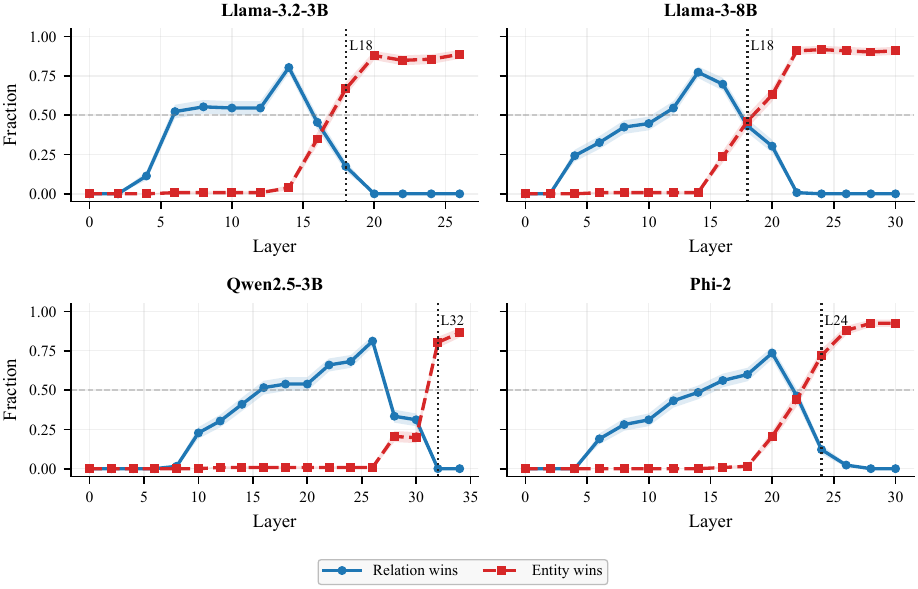}
  \caption{\textbf{Both-change competition (all four models).}
    \relc{Relation wins (blue)} dominate middle layers;
    \entc{entity wins (red)} dominate late layers.
    Dotted vertical lines = crossover layers, aligning
    with entity onset from Experiment~1.}
  \label{fig:both}
\end{figure}

\section{Steering Figure and Table}
\label{app:steer}

\begin{figure}[h]
  \centering
  \includegraphics[width=0.85\textwidth]{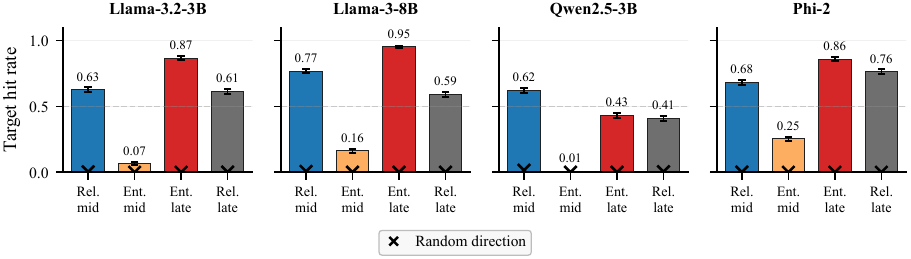}
  \caption{\textbf{Steering temporal asymmetry.}
    Relation directions (blue/orange bars) steer
    effectively in middle layers; entity directions
    (red bars) are substantially weaker mid-layer
    and strongest late. $\times$ marks near-zero
    random-direction baselines ($\leq 0.015$).
    Relation directions also remain effective late
    but are outcompeted by entity under direct
    conflict (Experiment~2).}
  \label{fig:steer}
\end{figure}

\begin{table}[!ht]
\centering
\caption{Steering hit rates at $\alpha = 1.0$. Random
  baselines $\leq 0.015$. R = relation, E = entity.
  Relation steers mid-layer; entity steers late.}
\label{tab:steer}
\vspace{2pt}
\setlength{\tabcolsep}{3pt}
\small\renewcommand{\arraystretch}{1.12}
\begin{tabular}{lcccc}
\toprule
Model & R@mid & E@mid & E@late & R@late \\
\midrule
Llama-3.2-3B & $0.626$ & $0.065$ & $0.867$ & $0.612$ \\
Llama-3-8B   & $0.768$ & $0.162$ & $0.952$ & $0.589$ \\
Qwen2.5-3B   & $0.620$ & $0.009$ & $0.430$ & $0.407$ \\
Phi-2        & $0.682$ & $0.253$ & $0.859$ & $0.764$ \\
\bottomrule
\end{tabular}
\end{table}

\clearpage

\section{Threshold Sensitivity}
\label{app:threshold}

\begin{table}[H]
\centering
\caption{Pair-balanced onset across thresholds $0.2$--$0.5$.
  Relation-before-entity holds in all 16
  model-threshold combinations ($\checkmark$).}
\small\renewcommand{\arraystretch}{1.08}
\setlength{\tabcolsep}{4pt}
\begin{tabular}{L{2.2cm}C{1.0cm}C{1.2cm}C{1.2cm}
                C{0.8cm}C{0.8cm}}
\toprule
\textbf{Model} & \textbf{Thresh.} &
\textbf{Rel.\ onset} & \textbf{Ent.\ onset} &
\textbf{Gap} & \textbf{Rel.$<$Ent.} \\
\midrule
Llama-3.2-3B & 0.2 & L6  & L16 & 10 & $\surd$ \\
Llama-3.2-3B & 0.3 & L6  & L18 & 12 & $\surd$ \\
Llama-3.2-3B & 0.4 & L6  & L18 & 12 & $\surd$ \\
Llama-3.2-3B & 0.5 & L14 & L18 &  4 & $\surd$ \\
\midrule
Llama-3-8B   & 0.2 & L4  & L18 & 14 & $\surd$ \\
Llama-3-8B   & 0.3 & L6  & L20 & 14 & $\surd$ \\
Llama-3-8B   & 0.4 & L10 & L20 & 10 & $\surd$ \\
Llama-3-8B   & 0.5 & L14 & L20 &  6 & $\surd$ \\
\midrule
Qwen2.5-3B   & 0.2 & L12 & L32 & 20 & $\surd$ \\
Qwen2.5-3B   & 0.3 & L14 & L32 & 18 & $\surd$ \\
Qwen2.5-3B   & 0.4 & L16 & L32 & 16 & $\surd$ \\
Qwen2.5-3B   & 0.5 & L22 & L32 & 10 & $\surd$ \\
\midrule
Phi-2        & 0.2 & L8  & L22 & 14 & $\surd$ \\
Phi-2        & 0.3 & L12 & L24 & 12 & $\surd$ \\
Phi-2        & 0.4 & L14 & L24 & 10 & $\surd$ \\
Phi-2        & 0.5 & L18 & L24 &  6 & $\surd$ \\
\bottomrule
\end{tabular}
\end{table}

\section{Additional Controls}
\label{app:controls}

\paragraph{Wrong-entity relation control.}
The wrong-entity control patches a different entity of
the same donor family, distinguishing two possible
outputs: \emph{relation-only transfer} (donor relation
applied to the recipient entity, e.g., output =
\textit{French}) versus \emph{donor-answer copying}
(donor's specific answer for the wrong entity).

Peak relation-only transfer is read from
\texttt{relation\_wrong\_entity\_summary.csv} by taking
the layer with maximum \texttt{relation\_only\_transfer\_pct}
per model:

\begin{table}[h]
\centering
\caption{Wrong-entity control: peak relation-only transfer
layer per model. Rel.-only denotes the donor relation applied
to the receiver entity; donor-copy denotes copying the donor's
specific answer. Donor-copy is near zero at these layers and
rises only later once entity commitment takes over.}
\label{tab:wrongentity}
\small\renewcommand{\arraystretch}{1.1}
\begin{tabular}{lcccc}
\toprule
\textbf{Model} & \textbf{Peak layer} &
\textbf{Rel.-only} & \textbf{Donor-copy} &
\textbf{Exp.~2 rel.\ peak} \\
\midrule
Llama-3.2-3B & L14 & 0.818 & 0.038 & L14 \\
Llama-3-8B   & L14 & 0.788 & 0.000 & L14 \\
Qwen2.5-3B   & L26 & 0.811 & 0.000 & L26 \\
Phi-2        & L20 & 0.864 & 0.114 & L20 \\
\bottomrule
\end{tabular}
\end{table}

The peak wrong-entity layer matches the peak relation-wins
layer from Experiment~2 in every model. This alignment
provides convergent evidence that the wrong-entity control
and both-change competition identify the same relation-dominant
middle-layer regime. At these layers relation-only transfer
reaches $0.79$--$0.86$ while donor-answer copying remains
$\leq 0.11$; donor-answer copying rises substantially only in
later layers once entity commitment takes over at the final token.
This rules out donor-answer copying as an explanation for
mid-layer relation transfer.

\paragraph{Both-change competition controls.}
Table~\ref{tab:both_controls} summarises the four control
conditions for Experiment~2. Noise max is the maximum
structured win rate (entity or relation) across all layers
under random Gaussian patching. Self orig.\ is the mean
original-retained rate under self-patch. Unrel.\ ent.\ is
peak entity-like overwrite from unrelated donors, which is
high in late layers when the final-token state becomes
broadly overwrite-sensitive. Unrel.\ rel.\ is peak
relation-wins from unrelated donors, which remains near
zero throughout.

\begin{table}[h]
\centering
\caption{Both-change control diagnostics. Noise patches
  produce near-zero structured wins; self-patches preserve
  the original output; unrelated donors show high late
  entity-like overwrite but near-zero relation-wins,
  indicating mid-layer relation dominance is not a
  generic patching artifact.}
\label{tab:both_controls}
\small\renewcommand{\arraystretch}{1.1}
\begin{tabular}{lcccc}
\toprule
\textbf{Model} & \textbf{Noise max} &
\textbf{Self orig.} &
\textbf{Unrel.\ ent.} &
\textbf{Unrel.\ rel.} \\
\midrule
Llama-3.2-3B & 0.000 & 1.000 & 0.970 & 0.015 \\
Llama-3-8B   & 0.008 & 1.000 & 0.924 & 0.030 \\
Qwen2.5-3B   & 0.008 & 0.992 & 0.917 & 0.008 \\
Phi-2        & 0.008 & 0.992 & 0.932 & 0.008 \\
\bottomrule
\end{tabular}
\end{table}

\paragraph{Unrelated-donor overwrite diagnostic.}
Unrelated-donor patches produce high entity-like overwrite
only in the late regime, marking broadly overwrite-sensitive
final-token states. However, relation-wins from unrelated
donors remain $\leq 0.030$, ruling out generic patching as
an explanation for mid-layer relation dominance.

\section{Prompt Family Details}
\label{app:families}

\begin{table}[H]
\centering
\caption{Prompt families, templates, and record counts
  used in the controlled prompt banks.}
\small\renewcommand{\arraystretch}{1.08}
\setlength{\tabcolsep}{4pt}
\begin{tabular}{L{2.2cm}L{4.2cm}C{0.8cm}}
\toprule
\textbf{Family} & \textbf{Template} & \textbf{N} \\
\midrule
capital       & The capital of \{X\} is        & 33 \\
language      & The official language of \{X\} is & 33 \\
past tense    & The past tense of \{X\} is     & 21 \\
present part. & The present participle of \{X\} is & 21 \\
plural        & The plural of \{X\} is         & 27 \\
opposite      & The opposite of \{X\} is       & 12 \\
comparative   & The comparative form of \{X\} is & 12 \\
symbol        & The chemical symbol for \{X\} is & 12 \\
\bottomrule
\end{tabular}
\end{table}

Prompt items are programmatically defined from fixed item
banks. We audit greedy generation and first-answer-token
statistics, including rank and logit margin, to identify
ambiguous or unstable items and to document prompt quality.

\section{Code and Reproducibility}
\begin{sloppypar}
Code and results available at
\url{https://github.com/divyanshddn146/deferred-entity-commitment}
\end{sloppypar}

\end{document}